\documentclass[letterpaper, 10 pt, conference]{ieeeconf}  

\IEEEoverridecommandlockouts                              

\usepackage{cite}
\usepackage{amsmath,amssymb,amsfonts}
\usepackage{graphicx}
\usepackage{textcomp}
\usepackage{xcolor}
\usepackage[ruled,linesnumbered, vlined]{algorithm2e}
\DontPrintSemicolon
\SetCommentSty{myCommentStyle}

\usepackage{amsfonts}
\let\labelindent\relax
\usepackage{enumitem}
\usepackage{amsmath}
\usepackage{graphicx}
\usepackage{subfig}
\usepackage[margin=0.792in]{geometry}
\usepackage{bbm}
\usepackage{algpseudocode} 
\usepackage{multirow}
\usepackage{xcolor}

\newcommand\Tstrut{\rule{0pt}{2.6ex}}         
\newcommand\Bstrut{\rule[-0.9ex]{0pt}{0pt}}   
\def\BibTeX{{\rm B\kern-.05em{\sc i\kern-.025em b}\kern-.08em
    T\kern-.1667em\lower.7ex\hbox{E}\kern-.125emX}}

\title{\LARGE \bf
Multi-Stage Monte Carlo Tree Search for Non-Monotone Object Rearrangement Planning in Narrow Confined Environments
}

\author{Hanwen Ren and Ahmed H. Qureshi
\thanks{*This work was supported by the National Science Foundation (NSF) under award no. 2204528.}
\thanks{Hanwen Ren and Ahmed H. Qureshi are with the Department of Computer Science, Purdue University, West Lafayette, IN, USA, 47907. Email {\tt\small$\{$ren221, ahqureshi$\}@$purdue.edu}}%
}

\begin{document}

\maketitle
\thispagestyle{empty}
\pagestyle{empty}

\begin{abstract}
Non-monotone object rearrangement planning in confined spaces such as cabinets and shelves is a widely occurring but challenging problem in robotics. Both the robot motion and the available regions for object relocation are highly constrained because of the limited space. This work proposes a Multi-Stage Monte Carlo Tree Search (MS-MCTS) method to solve non-monotone object rearrangement planning problems in confined spaces. Our approach decouples the complex problem into simpler subproblems using an object stage topology. A subgoal-focused tree expansion algorithm that jointly considers the high-level planning and the low-level robot motion is designed to reduce the search space and better guide the search process. By fitting the task into the MCTS paradigm, our method generates short object rearrangement sequences by balancing exploration and exploitation. The experiments demonstrate that our method outperforms the existing methods in terms of the planning time, the number of steps, the object moving distance and the gripper moving distance. Moreover, we deploy our MS-MCTS to a real-world robot system and verify its performance in different scenarios. 
\end{abstract}

\section{Introduction}
Object rearrangement planning in narrow, confined spaces such as cabinets, shelves, and fridges is essential for robots working in such environments. For example, robots must rearrange objects by grouping the same type for maintenance needs and create particular patterns to use the confined space better. Object rearrangement planning is generally known as NP-hard \cite{reif1994motion, wilfong1988motion} because the planner needs to figure out the moving order of the objects and the intermediate relocation regions. This problem can further be categorized as follows based on the moving count of objects and robot actions. The monotone instances, where each object can be relocated at most once, and the non-monotone instances, where they can be relocated multiple times in the scene. Regarding robot movements, the prehensile instances consider robot pick-and-place actions, whereas non-prehensile instances use push actions.\par
This work focuses on prehensile non-monotone object arrangement planning problems in narrow, confined spaces. The confined setting introduces an extra constraint on the robot's motion compared to tabletop environments. In tabletop environments, robots can grasp objects from the top and use the space above the objects to avoid collisions during relocation. However, in confined environments with a covered top and a side opening, the objects can only be accessed from the side. Hence, robots always occupy a certain amount of space during the motion, resulting in fewer regions for objects to be placed and a high chance of object-to-object and object-to-robot collisions. \par
\begin{figure}[t]
\includegraphics[trim={0.1cm 0 0 0cm},clip,width=8.5cm]{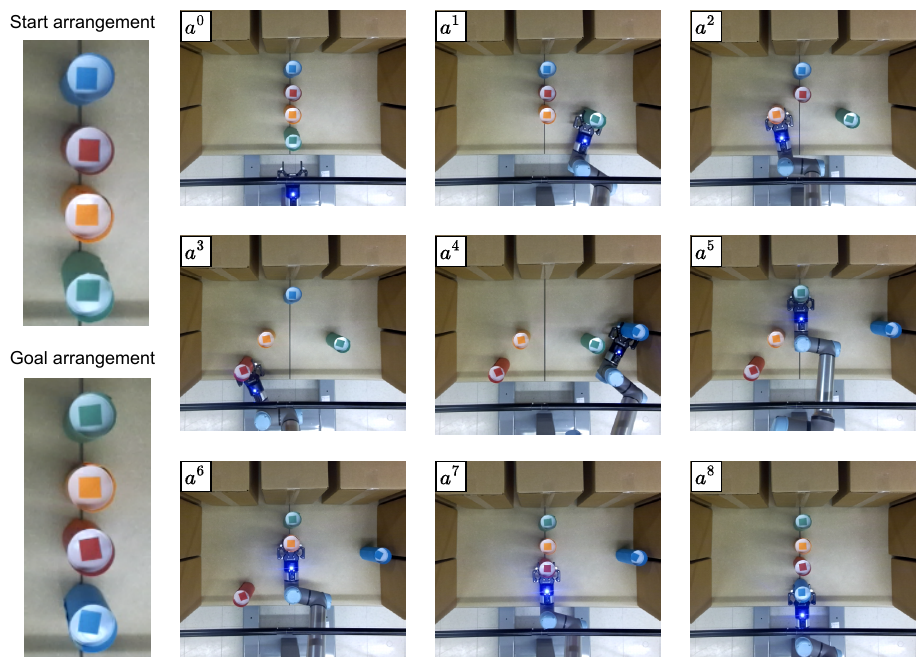}
\centering
\caption{The figure shows a four objects flipping case with complex object dependency relations. Our method solves it in 8 steps without redundant robot actions.}%
    \label{fig: figure 3}\vspace{-0.3in}
\end{figure}
Most existing works solve the non-monotone rearrangement planning problem using diverse tree search variations. The general method treats the start arrangement as the root, and the tree keeps growing until the goal arrangement is achieved. The parent and child tree nodes are linked with a single object movement, i.e., an object is relocated to another region by the robot in a collision-free manner. Once the goal arrangement is found, the algorithm backtracks and recovers the entire plan. Due to the elevated complexity of the non-monotone cases, intelligent search algorithms are developed to solve them in a reasonable time budget \cite{wang2022efficient}. 
However, most of these approaches only aim to find a feasible solution, while the quality of the result, such as the number of steps and distance traveled by the robot, are not optimized. 

Therefore, in this paper, we propose an efficient Multi-Stage Monte Carlo Tree Search (MS-MCTS) approach that solves prehensile non-monotone object rearrangement planning problems in narrow, confined spaces with a higher success rate and fewer steps and less distance traveled than any existing methods.
We also deploy the method to real-world scenarios and verify its sim-to-real generalization abilities. In summary, the main contributions of the proposed work are listed as follows:
\begin{itemize}
    \item A non-monotone object rearrangement planner that finds high-quality solutions by fitting the problem into the MCTS paradigm. Our method suits real-world robotics systems of different configurations.
    \item A specially designed subgoal-focused tree expansion algorithm that jointly considers the high-level relocation planning and low-level robot motion planning, which constrains the search within a limited sub-space.
    \item A novel object relocation order heuristic helps the planner decouple the complicated problem into simpler sub-problems, which is proven suitable for the narrow, confined spaces setting.
    \item A computationally efficient robot motion planner that minimizes the swept volume of the robot actions and further leads to higher chances of finding valid plans.
\end{itemize} 

\section{Related Work}
Object rearrangement planning in various environments is an active and widely researched problem in robotics, which is also a frequently occurring instance in the field of Navigation Among Movable Obstacles (NAMO) \cite{chen1990practical, stilman2005navigation} and Task And Motion Planning (TAMP) \cite{garrett2021integrated, srivastava2014combined}. Since it needs to consider both the high-level task planner and the low-level motion planner in environments with movable objects, the problem is generally considered NP-hard \cite{wilfong1988motion}. Existing works solve the rearrangement planning problem under different settings. \cite{stilman2007planning, stilman2008planning, stilman2007manipulation} solves monotone instances by checking all the permutations of the object relocation order in a reverse-time manner. \cite{krontiris2016efficiently, havur2014geometric, gao2022fast, liu2022structformer, curtis2022long, goodwin2022semantically} perform the planning on tabletop environments utilizing tree search with modified growing strategies or search hierarchy followed by backtracking. Others \cite{qureshi2021nerp, zeng2021transporter, yuan2019end, yuan2018rearrangement, goyal2022ifor} put the recent advancement of the deep neural network into play and let the planning agent learn the underlying logic of various moves using the collected dataset. Aside from the open workspace, \cite{wang2022efficient, wang2021uniform, wang2022lazy, lee2021tree} assume more constrained environments like the cabinets or other confined spaces with only one opening in the front, which are more common in real-world scenarios. Their strategies include performing intelligent expansions or pre-pruning the search trees so that invalid actions can be filtered out at early stages. These moves shrink the search space and increase efficiency. Compared with our method that finds solutions involving critical moves only, most of them aim to find a valid solution in a depth-first search manner without focusing on the quality of the resulting plans. \par
In order to find high-quality solutions regrading various objectives, the Monte Carlo Tree Search (MCTS) \cite{chaslot2008monte} has recently been applied to the rearrangement planning problems. As a best-first search method, MCTS does not require a positional evaluation function to guide the search. Instead, it explores the search space randomly and gradually becomes better at estimating the quality of the best move \cite{james2017analysis}. MCTS has been proven to be a general algorithm which requires little or no domain knowledge while it is extremely useful in balancing the exploration and exploitation during the search process, resulting in strong performances \cite{browne2012survey}. The AlphaGo, AlphaZero \cite{silver2018general, fu2016alphago, moerland2018a0c} took advantage of MCTS and beat the top-ranked human go player in 2016. In rearrangement planning problems, work \cite{labbe2020monte} use MCTS to achieve both efficiency and scalability in tabletop environments. In addition, others \cite{song2020multi, king2017unobservable} fit the non-prehensile object rearranging and sorting tasks into the MCTS paradigm to achieve decent performance. Different from these works, we use the MCTS in the confined spaces setting with specially designed algorithms to guide the expansion and simulation process, resulting in improved performances in the runtime and the plan quality.

\begin{figure*}[t]
    \centering
    \subfloat[\centering Linear motion planner]{{\includegraphics[trim={0.4cm 0cm 0.4cm 0cm},clip, width=0.365\textwidth]{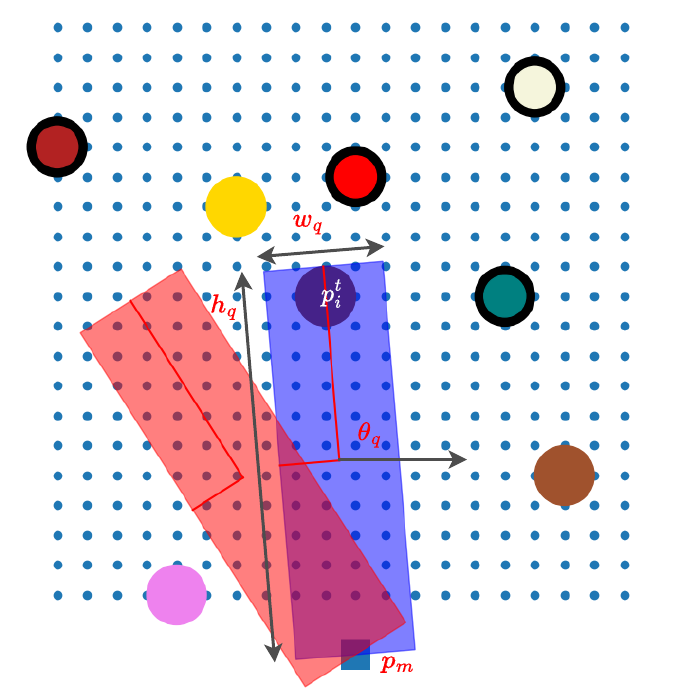} }}%
    \qquad
    \subfloat[\centering Object stage topology generation]{{\includegraphics[width=0.55\textwidth]{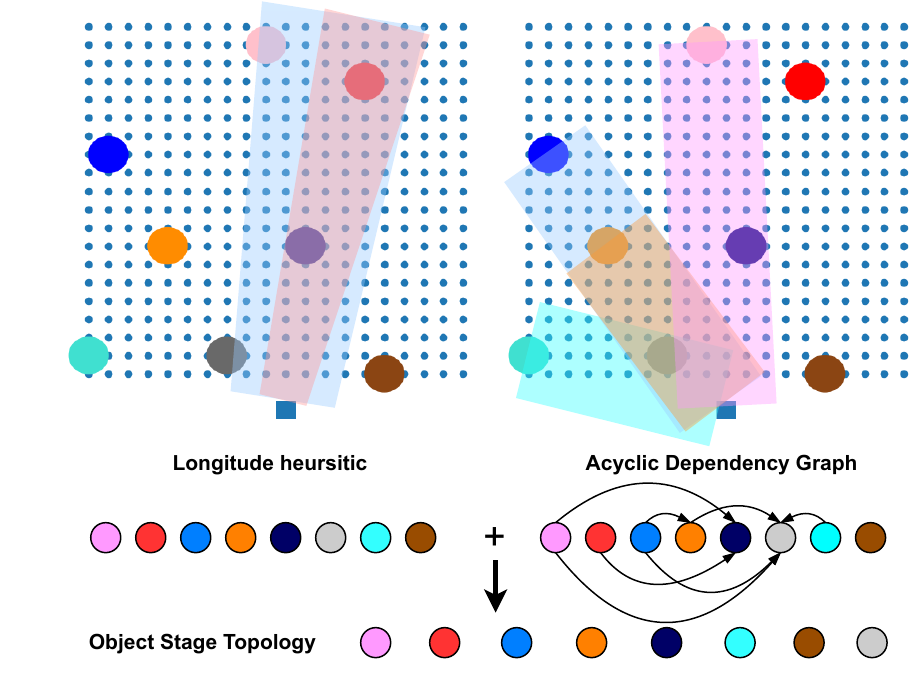} }}%
    \caption{(a) The gripper at $p_m$ uses the collision-free picking path (blue) and placing path (red) generated from our linear motion planner to relocate the object at $p^t_i$. Objects with a black contour are already placed at their goal regions. (b) The left sub-figure shows the order from the longitude heuristic while the right one illustrates the generation process of the dependency graph. For example, the goal region of the grey object blocks the turquoise, blue, orange, and pink objects, which creates four edges toward the grey object in the dependency graph. The final object stage topology is created by jointly considering the longitude heuristic and the dependency graph.}%
    \label{fig:eq and topo}%
    \vspace{-0.2in}
\end{figure*}

\section{Problem Formulation}
Let a confined workspace with one narrow opening at the front be denoted as $\mathcal{W} \subset \mathbb{R}^3$. In this confined space, $n \in \mathbb{N}$ identical but uniquely labeled cylindrical objects, with radius $b$, are denoted as $\mathcal{O} = \{o_1,\dots,o_n\}$. The ground surface of the workspace has dimension of $(d_x, d_y)$ in the $X, Y$ axes, respectively. Each object can be placed at a random location $p = (x_p, y_p) \in \mathcal{P} \subset \mathbb{R}^2$ on the workspace surface as long as it is collision-free.
The placement locations associated with all objects at step $t$ form the object arrangement $a^t = \{p^t_1,\dots,p^t_n\} \subset \mathcal{A}$, where $\mathcal{A}$ is the arrangement space. The $a^t[o_i] = p^t_i$ denotes that object $o_i$ locates at region $p^t_i$ in arrangement $a^t$. A robot arm $\mathcal{M}$ equipped with a gripper is placed in front of the workspace opening at location $p_m$ to perform prehensile object relocation actions. One relocation action $r^t = (r^t_{pick}, r^t_{place})$ at step $t$ involves both picking a certain object at its current region and placing it at the next region, resulting in a new object arrangement $a^{t+1}$. While performing such actions, the robot follows a certain manipulation path $\pi(r^t) = \{q^t_0,...,q^t_k\}$, where each $q_i \in \mathcal{Q} \subset \mathbb{R}^d$ is an instance in the $d$-dimensional arm configuration space. The volume occupied by the robot arm during a relocation action is represented as $V(\pi(r^t))$. A relocation action involving object $o_k$ at step $t$ is valid if it satisfies the collision constraint, written as $V(\pi(r^t)) \cap V(a^t \backslash a^t[o_k]) = \emptyset$, where the term, $V(a^t \backslash a^t[o_k])$, is the space occupied by all other objects except the one that is currently being relocated.\par
Using the notations above, the non-monotone object rearrangement planning problem is formally defined as follows. Given $n$-objects start arrangement $a^s$ and goal arrangement $a^g$, find a short valid robot action sequence R = $\{(r^0_{pick}, r^0_{place}), (r^1_{pick}, r^1_{place}),\cdots \}$ that relocates all objects from their start to the goal placements, minimizing the objects and the gripper moving distance.

\section{Methodology}
This section formally introduces the proposed Multi-Stage Monte-Carlo Tree Search (MS-MCTS) object rearrangement planner and the related modules in detail.
\subsection{Linear Motion Planner}
We design a Linear Motion Planner (LMP) to move the gripper toward the objects and perform prehensile actions. In the confined spaces setting, where the robot cannot use the space on the upper $Z$ axis to perform grasp action and avoid collisions, the linear planner minimizes the swept volume by first aiming the gripper toward the objects and then moving it linearly in the $XY$ plane while maintaining a fixed height on the $Z$ axis. The robot configurations for linear movement $\pi(r^t)$ can be obtained by calculating the Inverse Kinematics (IK) \cite{chan1987iterative} on the discretized points along the path. To relocate an object $o_i$ from its current region $p^t_i$ to the next region $p^{t+1}_i$, the gripper first moves from its home location $p_m$ to $p^t_i$, picks the object up, and retrieve it back. Then it goes to $p^{t+1}_i$, places the object, and returns to $p_m$. In order to check collisions, the robot's swept volume $V(\pi(r^t))$ during the linear movement is constructed by a rectangular tunnel with length $h_q$, width $w_q$, tilted at angle $\theta_q$ shown in Fig. \ref{fig:eq and topo} (a). In the rest of the paper, we use the terms picking path and placing path to denote the swept volume of the robot's prehensile actions. The rectangular path parameters $h_q$ and $\theta_q$ that moves the gripper from $p_m$ to an object with radius $b$ located at $p^t_i$ can be calculated by $h_q = \left\lVert \boldsymbol{p^t_i} - \boldsymbol{p_m} \right\rVert_2 + b$, $\theta_q = \arccos{\biggl(\frac{\boldsymbol{\hat{u}}\cdot (\boldsymbol{p^t_i} - \boldsymbol{p_m})}{\left\lVert \boldsymbol{p^t_i} - \boldsymbol{p_m} \right\rVert_2}\biggr)}$,
where $\boldsymbol{\hat{u}}$ is the unit vector on the positive $X$ axis. The parameter $w_q$ is the robot's maximum width when the gripper moves linearly. There are three advantages of our linear motion planner. First, it is a general method that is not specifically tied to the UR5e we use in the real world experiments. 
Second, since the robot always stays in the picking and placing path during its motion, the linear motion planner minimizes the swept volume, resulting in a higher probability of finding valid solutions. Third, the swept volume associated with the LMP can be easily calculated using the equations mentioned above and its collision with the objects can be efficiently checked by the Separating Axis Theorem \cite{lin1997collision}, which facilitates the searching speed and leads to a higher success rate in a fixed time budget. 

\subsection{Object Stage Topology Generation}
We create an object stage topology that defines the desirable order to place all objects in their goal regions, which has the highest chance of solving the instance efficiently by considering the longitude heuristic and the dependency graph of goal arrangement jointly. The longitude heuristic prioritizes objects that are to be placed farther behind than closer to the robot. Recall the linear motion planner introduced above. The objects placed at their goal regions in the front side of the environment are more likely to block the placing paths for objects further back as shown in the left image of Fig. \ref{fig:eq and topo} (b). This implies that objects with front-end goals must be relocated to an additional buffer region before returning, wasting the steps of getting them to the goal regions in the first place. Thus, it is more appropriate to relocate objects in the decreasing order of their longitudinal distances between their goals and the robot. Apart from the longitude heuristic, the desirable object order should also obey the acyclic dependency graph of the goal arrangement. The dependency graph is created by checking if the goal region of each object $o_i$ prevents other objects from being placed at their goal regions, if so, there will be edges going from the objects that are blocked by $o_i$. Object $o_i$ should not be placed at its goal region when there are still edges points to it. The dependency graph is created based on the underlying motion planner, for our LMP, one example is illustrated as the right image of Fig. \ref{fig:eq and topo} (b). The final object stage topology is generated by performing the topological sort on the dependency graph while respecting the longitude heuristic for objects that are not dependent on each other.
\SetInd{0.25em}{0.5em}
\begin{algorithm}[hbt!]
\caption{Feasible Buffer Region Generation}\label{alg:propose_new_regions}
\KwData{$o_i, o_k, \{o_d\}, a^t, a^g$ }
\KwResult{$P^r_i = \{p_i^1, \dots, p_i^m\}$}
$P^r_i = \emptyset$\;
$V(\pi(r_{pick})) = \text{MP}(a^t[o_k])$\;
$V(\pi(r_{place})) = \text{MP}(a^g[o_k])$\;
$\{V(\pi(r^d_{pick}))\} = \text{MP}(a^t[\{o_d\}])$ \;
\For{\textup{region candidates} $p_i \in P$}{
    \If{\textup{collision\_free}$(p_i, \{a^t\backslash a^t[o_i], \{V(\pi(r^d_{pick}))\},$ \; 
    ~~~~~~~~~~~~~~~~~~~~~~~~$V(\pi(r_{pick})), V(\pi(r_{place}))\})$}{
        $V(\pi(r^i_{place})) = \text{MP}(p_i)$\;
        \If{\textup{collision\_free}$(V(\pi(r^i_{place})), a^t\backslash a^t[o_i])$}{
            $P^r_i\textup{.add}(p_i)$\;
            \If{\textup{size}$(P^r_i) = m$}{
                \textbf{break}
            }
        }
    }
}
\Return $P^r_i$
\end{algorithm}
\subsection{Single-Stage MCTS planner}
Our Single-Stage MCTS (SS-MCTS) planner aims to move one specific object $o_k$ from its current region $a^t[o_k]$ to the goal $a^g[o_k]$. In other words, the search halts when the objective $a^{t'}[o_k] = a^g[o_k]$ is achieved at some step $t' \geq t$. To better present our ideas, we introduce several new notations as follows. Assume we are now in the SS-MCTS planner that focuses on the $k$-th object $o_k$ in the generated topology, and the current step count is $t$. All the objects that have been relocated to their goal regions form a set of static objects $\mathcal{O}_s= \{o_{1},\dots, o_{k-1}\}$ and the rest are represented as $\mathcal{O}_r= \{o_{k},\dots, o_{n}\}$. In the SS-MCTS focusing on object $o_k$, only the objects in $\mathcal{O}_r$ are subject to be relocated. The current arrangement $a^t$ and the goal arrangement $a^g$ contain the most up-to-date information about the 
objects' current/goal regions. Inside the tree, the parent and child tree nodes are linked by robot action $r^t$ that relocates a single object $o_i$ in $\mathcal{O}_r$ from region $p^t_i$ to $p^{t+1}_i$, forming a new arrangement $a^{t+1}$. The associated reward function value is assigned as the negation of the Euclidean distance between two regions, i.e., $-\|p^t_k - p^{t+1}_k\|_2$. The following paragraphs reveal the details of our SS-MCTS by fitting them into the standard MCTS paradigm.\par
\textbf{Selection}: The SS-MCTS planner balances the exploration and exploitation by utilizing the tuned Upper Confidence Bound (UCB) \cite{browne2012survey}.
In the selection process, we pick a leaf node by going through successive child nodes maximizing the UCB values starting from the root. If the selected leaf node has already been visited, the expansion module introduced below is applied to further grow the tree.
\setlength{\textfloatsep}{0pt}

\SetInd{0.25em}{0.5em}
\begin{algorithm}[bt!]
\caption{Expansion}\label{alg:SS-MCTS expansion}
\KwData{TreeNode $T_p(o_k, a^t, a^g, O_r)$\;}
\KwResult{TreeNode $T_c(o_k, a^{t+1}, a^g, O_r)$\;}
$O_b = \textup{get\_blocking\_objects}(o_k, a^t, a^g, O_r)$\;
\uIf{$O_b \neq \emptyset$}{
    \While{True}{
        $O'_b = \emptyset$\;
        \For{$o_i \in O_b$}{
            $V(\pi(r^i_{pick})) = \text{MP}(a^t[o_i])$\;
            $O^{ip}_b = \textup{collision\_objs}(O_r\backslash o_i, V(\pi(r^i_{pick})))$\;
            \uIf{$O^{ip}_b = \emptyset$ }{
                    \For{$q$ $\leftarrow$ $i-1$ \text{to} $k+1$}{
                        $P^r_i = \textup{new\_region}\;(o_i, o_k,  o_{\{k+1 \dots q\}}, a^t, a^g$)\;
                        \If{$P^r_i$}{
                        \For{$p^r_i \in P^r_i$}{
                            $T_p.\textup{add\_child}(T_c(a^{t+1}[o_i] = p^r_i]))$\;
                        }
                        \textbf{break}}}
                
            }
            \Else{
                \For{$o_j \in O^{ip}_b$}{
                    $V(\pi(r^j_{pick})) = \textup{MP}(a^t[o_j])$\;
                    \uIf{\textup{collision\_free}$(O_r\backslash o_j, V(\pi(r^j_{pick})))$}{
                       
                        \For{$q$ $\leftarrow$ $j-1$ \text{to} $k+1$}{
                        $P^r_j = \textup{new\_region}\;(o_j, o_k, o_{\{k+1\dots q, i\}}, a^t, a^g)$\;
                        \If{$P^r_j$}{
                        \For{$p^r_j \in P^r_j$}{
                            $T_p.\textup{add\_child}(T_c(a^{t+1}[o_j] = p^r_j]))$\;
                        }
                        \textbf{break}
                        }
                        }
                    }
                    \Else{
                        $O'_b$.add$(o_j)$
                    }
                }
            }
        }
        halting$\_$condition()
    }
}
\Else{
     $T_p.\textup{add\_child}(T_c(a^{t+1}[o_k] = a^g[o_k]))$
}
\Return $T_p.\textup{child}[0]$
\end{algorithm}

\textbf{Expansion}: We present an efficient subgoal-focused expansion algorithm that shrinks the search space, which in result leads to a higher chance of finding valid solutions. The objective of the SS-MCTS is carried over to be the subgoal of the expansion module, i.e., relocating object of interest $o_k$ to its goal region, and the tree expands focusing on it. Note that this task planning method suits all underlying motion planning methods. Thus, in the following explanation and pseudocode, we use the notation MP for the motion planner. In order to achieve the subgoal, first, the planner finds all objects $O_b$ currently preventing $o_k$ from being relocated to its goal region using the picking and placing paths based on the current and goal region of $o_k$. All remaining objects except $o_k$ should be moved to other regions if they are currently inside the two paths. In addition, because $o_k$ becomes a static object in the following SS-MCTS focusing on {other objects}, the planner needs to ensure the remaining objects can still be accessed after placing $o_k$ at its goal region. Thus, objects whose picking path intersects with $o_k$ placed at its goal region are added to the blocking object list $O_b$ as well.
Next, we propose Algorithm \ref{alg:propose_new_regions} to generate feasible buffer region candidates in the continuous space for the objects $o_i \in O_b$. The potential region candidate set $P$ (line 5) is generated by adding 2D Gaussian offsets with zero mean and diagonal covariance matrix to the regions in discreteized workspace. In our tests, the diagonal value in the covariance matrix and the workspace discreteization unit are both set to be the radius of the cylindrical objects $b$. A helper function called collision$\_$free() detects whether there is a collision between the inputs. Each valid region candidate $p_i = (x_{p_i}, y_{p_i}) \subseteq \mathbb{R}^2$ for object $o_i$ should not collide with the following elements: 1. all the regions currently occupied by other objects $a^t\backslash a^t[o_i]$. 2. the picking and placing paths regarding object of interest $o_k$. 3. the picking paths of a dependency objects set \{$o_d$\} (line 6). The dependency objects are those who need to be relocated to fulfill the subgoal but it is currently being blocked by $o_i$. 4. Finally, all region candidates that meet the abovementioned constraints form the final list of valid regions $P_i^r$ if their corresponding relocation paths are feasible (line 9). During the actual execution, we sort the potential region candidates $P$ by the increasing distance to $o_i$ so that it better aligns with the reward function. In addition, an upper tree expansion threshold is set to prevent the tree from growing too wide while still maintaining decent performance.

The complete tree expansion algorithm is shown in Algorithm \ref{alg:SS-MCTS expansion} with the aid of an additional helper function. The collision$\_$objs() function (line 7) takes a list of objects and a robot motion path as inputs and returns all objects in the list that collide with the input path. Also, the notation $O_b^{ip}$ and $O_b^{id}$ denote all objects that block the picking and placing paths of object $o_i$, respectively. The algorithm starts at parent tree node $T_p$ by checking if both the picking and placing paths for the object of interest $o_k$ are collision-free, if so, only one child tree node $T_c$ that puts $o_k$ directly to its goal region is created, which means the subgoal is achieved (line 1, 2, 29). Otherwise, the subgoal-blocking objects $o_i \in O_b$ are further divided into the following two categories where different strategies are applied. 1. If only the picking path for $o_i$ is available, valid regions that do not block the picking path for objects earlier in the topology are proposed using Algorithm \ref{alg:propose_new_regions} and added to the tree (line 9-14). 2. If even the picking path for $o_i$ is not available, the algorithm first finds all objects $O^{ip}_b$ blocking it. For each object $o_j \in O^{ip}_b$ that can be accessed, new regions are proposed with an additional dependency object setting as $o_i$ so that the picking path of $o_i$ can be cleared (line 16-24). On the other hand, objects that are not accessible will be added to $O_b'$ as the focus of the next iteration (line 26). In the halting$\_$condition() function, the expansion process stops if new tree nodes are created, otherwise, the subgoal-blocking object set $O_b$ will be replaced by $O_b'$ and the expansion enters the next round. However, if the search depth reaches a certain threshold without making progress, $O_b$ is changed to all the feasible objects, aiming to guide the search out of the stuck node. Finally, the expansion algorithm returns the first child node found during the process, from where the simulation starts (line 30). The high-level ideology of our expansion algorithm is that it grows the tree by hierarchically moving blocking objects out of the picking and placing paths for object of interest $o_k$. Eventually, both paths of $o_k$ become collision-free, and our subgoal is fulfilled. Furthermore, our method can handle complex non-monotone problems mentioned in work \cite{krontiris2015dealing} where $o_k$ must be relocated to a buffer region before others because $o_j \in O_b^{ip}$ can potentially be set as $o_k$ during the search process (line 16).\par
\textbf{Simulation}: This part follows the standard MCTS pipeline \cite{browne2012survey}. A rollout process grows a pathological tree from the chosen node in the selection or expansion module until the object of interest $o_k$ is relocated to its goal region. During the process, the relocation region for the object is randomly selected from the result returned by Algorithm \ref{alg:propose_new_regions}.\par
\textbf{Back-propagation}: We set the reward function to be the negation of the accumulated Euclidean object relocation distance when the simulation ends. All tree nodes from the leaf where the simulation starts until the root receive the reward and one visited count during the backpropagation process.\par
The SS-MCTS planner halts when the current tree node achieves the objective. The sub-plan $\{r^t,\dots, r^{t+q}\}$ can be recovered by backward tree traversal.

\begin{table*}[!ht]
  \fontsize{7}{5}\selectfont
  \begin{center}
    \begin{tabular}{c c c c c c}
      \hline
      \multirow{2}{*}{\textbf{Rearrangement Task planner}} & \multicolumn{5}{c}{\textbf{Easy $\&$ Medium cases}}\Tstrut\Bstrut \\
       & Success rate $(\%)$ $\uparrow$ \Tstrut\Bstrut & Planning time (s) $\downarrow$ & Number of steps $\downarrow$ & Relocation distance $\downarrow$ & Gripper distance $\downarrow$\\
      \hline
     BiRRT(mRS)\Tstrut\Bstrut & 97.92 & 5.55$\pm$11.22 & 19.25$\pm$10.53 & 224.23 $\pm$ 113.17 & 504.38 $\pm$ 263.43\\
     PERTS(CIRS)\Tstrut\Bstrut & 80.42 & 1.15 $\pm$ 5.62 & 36.54 $\pm$ 67.97 & 410.95 $\pm$ 754.01 & 856.88 $\pm$ 1574.2\\
     MS-MCTS (ours)\Tstrut\Bstrut & \textbf{100.00} & \textbf{1.09$\pm$1.46} & \textbf{8.04$\pm$2.47}  & \textbf{86.92$\pm$32.32} & \textbf{198.18$\pm$65.53}\\
      \hline
         ~\\
    \end{tabular}

    \begin{tabular}{c c c c c c}
      \multirow{2}{*}{\textbf{Rearrangement Task planner}} & \multicolumn{5}{c}{\textbf{Hard cases}}\Tstrut\Bstrut \\
       & Success rate $(\%)$ $\uparrow$ \Tstrut\Bstrut & Planning time (s) $\downarrow$ & Number of steps $\downarrow$ & Relocation distance $\downarrow$ & Gripper distance $\downarrow$\\
      \hline
     BiRRT(mRS)\Tstrut\Bstrut & 4.38 & 28.99$\pm$18.22 & 22.00$\pm$7.87 & 273.58$\pm$88.19 & 583.93$\pm$190.17\\
     PERTS(CIRS)\Tstrut\Bstrut & 44.38 & \textbf{3.87$\pm$7.09} & 73.44$\pm$61.51 & 810.25$\pm$673.95 & 1682.51$\pm$1398.51\\
     MS-MCTS (ours)\Tstrut\Bstrut & \textbf{100.00} & 5.29$\pm$7.04 & \textbf{14.65$\pm$2.54} & \textbf{156.86$\pm$41.08} & \textbf{355.43$\pm$77.01}\\
      \hline
    \end{tabular}
    \caption{Experiments results reflect that our MS-MCTS outperforms the baseline methods by a large margin.}
    \label{tab:table3}
  \end{center}
\vspace{-0.3in}
\end{table*}

\subsection{Multi-Stage MCTS planner $\&$ Post Optimization}
The MS-MCTS planner comprises $n$ ordered SS-MCTS. The SS-MCTS at index $i$ sets its object of interest as $o_i$ from the generated object stage topology introduced above. By extracting and combining all the sub-plans from the SS-MCTS planners, we get the initial global plan $R' = \{(r^0_{pick}, r^0_{place}), (r^1_{pick}, r^1_{place}),\cdots \}$ that moves all objects from the start arrangement $a^s$ to the goal arrangement $a^g$. The goal plan $R'$ then goes through two post optimization steps. In the first step, continuous action segments involving the same object $\{(r^t_{pick}, r^t_{place}),\cdots, (r^{t+k}_{pick}, r^{t+k}_{place})\}$ will be combined to one single action $\{(r^t_{pick}, r^{t+k}_{place})\}$. The second step checks all non-adjacent action pairs $(r^t,r^{t+k})$ concerning the same object and changes it to $(r^t = (r^t_{pick}, r^{t+k}_{place}))$ if the actions in between $\{r^{t},\cdots,r^{t+k}\}$ can be still be performed without collision. These two optimization steps shrink the length of the original plan and result in the final global plan $R$. In addition, for each robot action $r^t$, instead of strictly following the rectangular path used in the planning process, the robot figures out the minimal object retrieval distance in the picking path before rotating the base and sending it to the desired region along the placing path. This design makes the plan execution faster and more intelligent.

\begin{figure}[t]
\includegraphics[trim={2.5cm 0 1.5cm 2cm},clip, width=9cm]{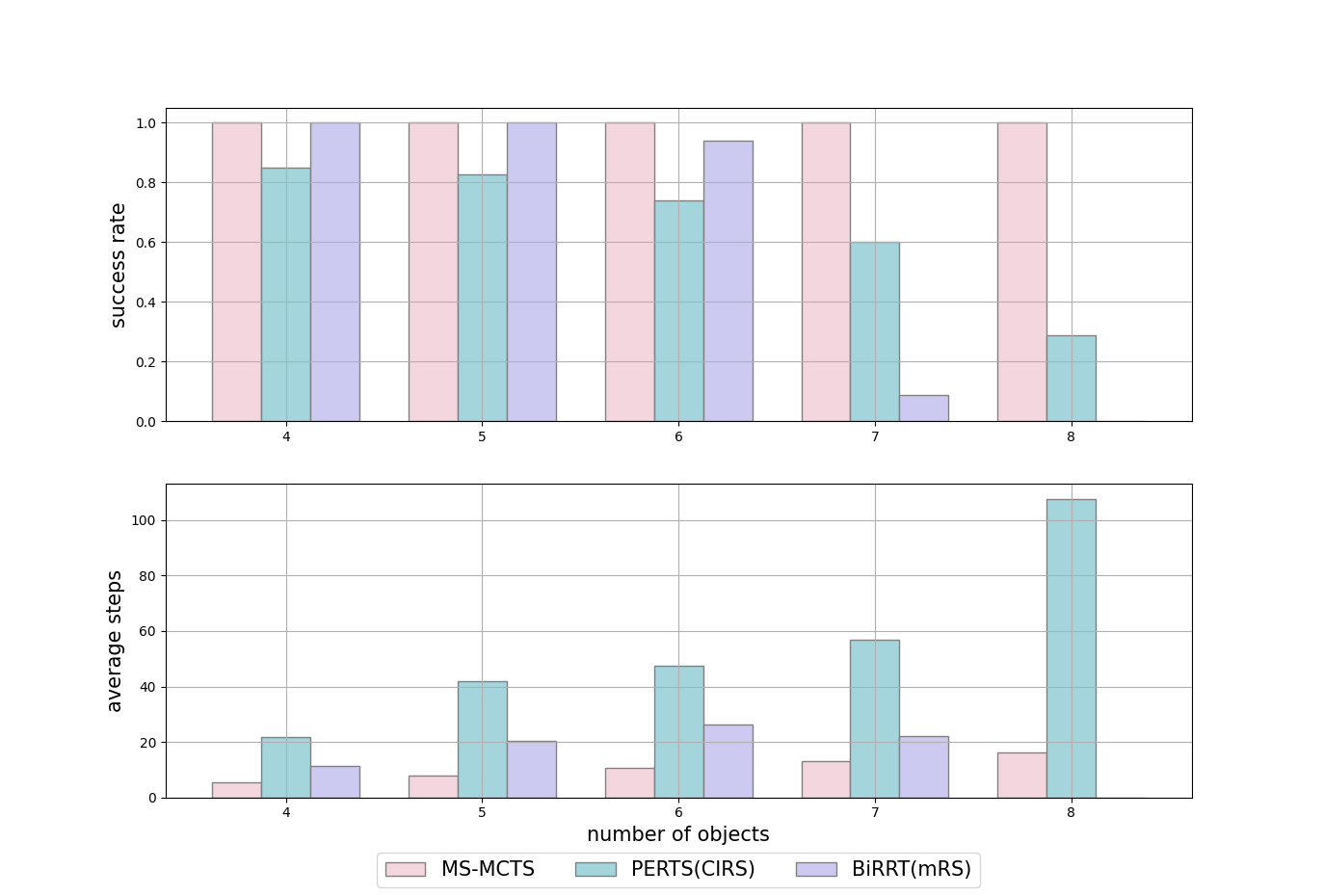}
\centering
\caption{Our MS-MCTS method has higher success rates and fewer steps than the baselines across all difficulty levels.}%
    \label{fig: figure 4}\vspace{0.1in}
\end{figure}

\section{Experiments}
\subsection{Simulation Setup}
We create randomly generated start and goal arrangements of five increasing difficulty levels to examine the performance of our MS-MCTS. The surface of the environment has a dimension of 20 by 20 units with 4-8 objects randomly placed inside. All objects are presented by circles with a radius 1 unit and they are placed with a minimal distance of 4 units between the centers to leave some grasping space for the gripper. The picking and placing rectangular path width of our linear motion planner is set to be 4 units. One example depicting the testing configurations is shown in Fig. \ref{fig:eq and topo} (b). The 4 objects cases are considered to be easy, 5-6 to be medium, and 7-8 to be hard. For each level involving different number of objects, 80 valid instances are randomly generated to test the method's performance comprehensively. Aside from our MS-MCTS, two additional prevailing non-monotone planner baselines, BiRRT(mRS) \cite{stilman2007manipulation, krontiris2016efficiently} and PERTS(CIRS) \cite{wang2022efficient}, are implemented to compare against our method. BiRRT(mRS) follows the BiRRT algorithm with tree nodes denoting different arrangements. The mRS monotone solver connects different tree nodes and finally forms the complete plan. PERTS(CIRS) divides the hard non-monotone case into a sequence of monotone cases. It utilizes perturbations to find valid buffer regions when the tree cannot grow further with the CIRS monotone solver. For all the methods, the underlying motion planner is set to be the LMP we introduced in the method section to increase the probability of finding valid solutions and enforce fairness. To evaluate the performance, we use the following metrics for quantitative evaluation:
\begin{itemize}[leftmargin=*]
    \item \textbf{Success rate}: Success rate tracks the percentage of successfully solved instances. Cases that are not solved within a 60 seconds time budget are considered as failures.
    \item \textbf{Planning time}: The planning time stores the time consumption for the task planner to return a solution.
    \item \textbf{Number of steps}: The number of steps measures the amount of object relocation actions needed to achieve the goal arrangement.
    \item \textbf{Relocation distance}: The relocation distance records the sum of the length that all the objects moved by the gripper.
    \item \textbf{Gripper distance}: The gripper distance calculates the sum of the length that the gripper moves to relocate all the objects to finish the task.
\end{itemize}
\begin{table*}[!ht]
  \fontsize{7}{5}\selectfont
  \begin{center}
    \begin{tabular}{c c c c c c}
      \hline
      \multirow{2}{*}{\textbf{Rearrangement Task planner}} & \multicolumn{5}{c}{\textbf{Easy $\&$ Medium cases}}\Tstrut\Bstrut \\
       & Success rate $(\%)$ $\uparrow$ \Tstrut\Bstrut & Planning time (s) $\downarrow$ & Number of steps $\downarrow$ & Relocation distance $\downarrow$ & Gripper distance $\downarrow$\\
      \hline
     MS-MCTS \Tstrut\Bstrut & 100.00 & 1.09$\pm$1.46 & 8.04$\pm$2.47  & 86.92$\pm$32.32 & 198.18$\pm$65.53\\
     MS-MCTS w/ Discrete buffer region\Tstrut\Bstrut & 100.00 & 1.48$\pm$2.4 & 7.98$\pm$2.39 & 87.4$\pm$31.96 & 198.21$\pm$63.82\\
     MS-MCTS w/o Object topo order\Tstrut\Bstrut & 88.75 & 1.8$\pm$4.98 & 7.95$\pm$2.5 & 96.93$\pm$38.39 & 215.72$\pm$77.51\\
     MS-MCTS w/o Post optimization\Tstrut\Bstrut & 100.00 & 0.64$\pm$0.91 & 8.68$\pm$2.54 & 90.45$\pm$32.63 & 208.43$\pm$66.07\\
      \hline
         ~\\
    \end{tabular}

    \begin{tabular}{c c c c c c}
      \multirow{2}{*}{\textbf{Rearrangement Task planner}} & \multicolumn{5}{c}{\textbf{Hard cases}}\Tstrut\Bstrut \\
       & Success rate $(\%)$ $\uparrow$ \Tstrut\Bstrut & Planning time (s) $\downarrow$ & Number of steps $\downarrow$ & Relocation distance $\downarrow$ & Gripper distance $\downarrow$\\
      \hline
     MS-MCTS \Tstrut\Bstrut & 100.00 & 5.29$\pm$7.04 & 14.65$\pm$2.54 & 156.86$\pm$41.08 & 355.43$\pm$77.01\\
     MS-MCTS w/ Discrete buffer region\Tstrut\Bstrut & 98.75 & 6.54$\pm$8.30 & 14.59$\pm$2.56 & 157.41$\pm$37.77 & 356.54$\pm$75.09\\
     MS-MCTS w/o Object topo order\Tstrut\Bstrut & 57.50 & 9.38$\pm$12.58 & 14.62$\pm$2.59 & 179.35$\pm$42.94 & 397.84$\pm$82.27\\
     MS-MCTS w/o Post optimization\Tstrut\Bstrut & 100.00 & 4.21$\pm$7.62 & 15.5$\pm$2.73 & 161.77$\pm$38.38 & 372.29$\pm$75.26\\
      \hline
    \end{tabular}
    \caption{Ablation study results reveal the impact of the continuous buffer region proposal, the object stage topology order, and the post optimization on the performance. We can see that all the abovementioned design choice helps to boost the overall result quality.}
    \label{tab:table2}
  \end{center}
\vspace{-0.3in}
\end{table*}
\subsection{Simulation Results}
Detailed visual comparisons of the success rate and the number of steps are shown in Fig. \ref{fig: figure 4} while Table \ref{tab:table3} lists all the quantitative results. All the metric items other than the success rate are averaged across all successfully solved testing instances. In the table, we separate the difficulty levels to better showcase the strong performance of our method. Under the easy plus medium settings, our method beasts the baselines in all metrics. With a $100\%$ success rate, our approach is fast in planning while still maintaining a decent solution quality that is at least three times better than the other methods in terms of the number of steps, the object relocation distance, and the gripper moving distance. When it comes to the difficult setting, our MS-MCTS still succeeded in solving all the instances due to the object topology order which shrinks the search space significantly by decoupling the complex problem into simple pieces while the next-best planner can only handle less than half of them. In addition, the subgoal-focused strategy allows our method to make only necessary moves to achieve the goal, which can be demonstrated by the low value in the last three metric items. The PERTS baseline beats us in the planning time because it is an average of the relatively manageable cases. The time consumption for the hardest cases is never recorded for PERTS as it fails in most of them, which can be seen from Fig. \ref{fig: figure 4}. In all, our approach achieves the highest success rate, the lowest time consumption, and the best result quality. Compared with the sampling-based methods used in the other two baselines, our subgoal-focused planning strategy only proposes necessary moves to fulfill the objective. As a result, the results generated from our method not only have the lowest time and execution consumption but also have a relatively low standard deviation in the same difficulty level. 
\vspace{-0.2in}
\subsection{Ablation Studies}
We conduct ablation studies to justify the design choice of our continuous buffer region proposal, the object stage topology order, and the post optimization. The results are summarized in Table \ref{tab:table2}. Since each of the designs has its pros and cons, we do not highlight the best performance of the metrics in the table.
\subsubsection{Continuous buffer region proposal}The success rate of the hard cases drops a little when we use the discrete region proposal using the object radius as the discretization unit. The planning time and the moving distance also increased by a small margin. In contrast, the continuous space proposal allows the planner to explore all the possible relocation regions, leading to high chances of finding optimal results in terms of total object relocation distance. Thus, choosing the continuous buffer region proposal is a better choice. 
\subsubsection{Object stage topology order} We replace our object order with a randomly generated counterpart. Without our object order, the success rate drops while the planning time increases significantly, especially in hard cases. Even considering the solved cases only, the object and gripper moving distance is worse than our complete algorithm. These results confirm that our object order is necessary for solving the complex object rearrangement tasks.
\subsubsection{Post optimization} From the table, we can see that the number of steps, as well as the relocation/gripper moving distance increase without this module. Although the post optimization adds slight time consumption, it improves the overall quality of the resulting task plans.

\subsection{Real Robot Experiments}
We deploy our MS-MCTS method on a UR5e robot arm manipulator equipped with a Robotiq 2F-85 gripper to solve various rearrangement planning problems in a confined environment with maximum dimensions of (140 cm, 70 cm, 38 cm). To better observe the robot motion from a top-down view, the testing space is constructed on a table surrounded by cardboard blocks on the three sides and a transparent "ceiling". The front end of the ceiling is represented by a bar attached to the top of both sides. This leaves one opening at the front for the robot to access the objects inside. The objects in the scene are cylindrical tubes with colored coating to help distinguish between them.

We set up four real-world scenarios of two medium and two hard configurations with structured patterns in goal arrangements. The medium cases contain an average of 5 objects, while the hard ones have an average of 8.5 objects. Our MS-MCTS planner takes $a^s,a^g$ as inputs, figures out the plan, and sends it to the robot arm manipulator for execution. During the experiment, our planner succeeded in finding valid plans for all test scenarios within 10 seconds. In the medium cases, the average number of steps is 8, and the average gripper moving distance is 5.56 m. While for the hard instances, the average number of steps is 16.5, and the average gripper moving distance is 11.45 m.
Fig. \ref{fig: figure 3} shows a successfully performed medium level four objects flipping case within a small workspace. We can clearly observe that the planner only generates the necessary moves to fulfill the goal arrangement. Also, the objects are relocated to a buffer region that is close to their current locations in order to minimize the gripper moving distance. The experiments of other scenarios are available in our supplementary videos.
\section{Conclusion}
This paper presents our Multi-Stage MCTS algorithm that solves non-monotone object rearrangement planning tasks in narrow, confined spaces. Our method decouples the generally considered NP-hard problems into a sequence of ordered stages, with each one focusing only on a specific object, which reduces the search space by a considerable amount. During relocation actions, the use of a linear motion planner minimizes the swept volume in the limited space and further leads to a higher chance of finding valid solutions. We fit the problem into the MCTS paradigm with customized designed functions to achieve high-quality results. The performance of our method is verified on various simulation cases with diverse difficulty levels and on the real robot. 
For future works, we seek to extend the planner into the 3D space with unknown, arbitrary objects to make it even more practical for real-world deployment.

\bibliographystyle{IEEEtran}
\bibliography{root}

\begin{thebibliography}{10}
\providecommand{\url}[1]{#1}
\csname url@samestyle\endcsname
\providecommand{\newblock}{\relax}
\providecommand{\bibinfo}[2]{#2}
\providecommand{\BIBentrySTDinterwordspacing}{\spaceskip=0pt\relax}
\providecommand{\BIBentryALTinterwordstretchfactor}{4}
\providecommand{\BIBentryALTinterwordspacing}{\spaceskip=\fontdimen2\font plus
\BIBentryALTinterwordstretchfactor\fontdimen3\font minus \fontdimen4\font\relax}
\providecommand{\BIBforeignlanguage}[2]{{%
\expandafter\ifx\csname l@#1\endcsname\relax
\typeout{** WARNING: IEEEtran.bst: No hyphenation pattern has been}%
\typeout{** loaded for the language `#1'. Using the pattern for}%
\typeout{** the default language instead.}%
\else
\language=\csname l@#1\endcsname
\fi
#2}}
\providecommand{\BIBdecl}{\relax}
\BIBdecl

\bibitem{reif1994motion}
J.~Reif and M.~Sharir, ``Motion planning in the presence of moving obstacles,'' \emph{Journal of the ACM (JACM)}, vol.~41, no.~4, pp. 764--790, 1994.

\bibitem{wilfong1988motion}
G.~Wilfong, ``Motion planning in the presence of movable obstacles,'' in \emph{Proceedings of the fourth annual symposium on Computational geometry}, 1988, pp. 279--288.

\bibitem{wang2022efficient}
R.~Wang, Y.~Miao, and K.~E. Bekris, ``Efficient and high-quality prehensile rearrangement in cluttered and confined spaces,'' in \emph{2022 International Conference on Robotics and Automation (ICRA)}.\hskip 1em plus 0.5em minus 0.4em\relax IEEE, 2022, pp. 1968--1975.

\bibitem{chen1990practical}
P.~C. Chen and Y.~K. Hwang, ``Practical path planning among movable obstacles,'' Sandia National Labs., Albuquerque, NM (USA), Tech. Rep., 1990.

\bibitem{stilman2005navigation}
M.~Stilman and J.~J. Kuffner, ``Navigation among movable obstacles: Real-time reasoning in complex environments,'' \emph{International Journal of Humanoid Robotics}, vol.~2, no.~04, pp. 479--503, 2005.

\bibitem{garrett2021integrated}
C.~R. Garrett, R.~Chitnis, R.~Holladay, B.~Kim, T.~Silver, L.~P. Kaelbling, and T.~Lozano-P{\'e}rez, ``Integrated task and motion planning,'' \emph{Annual review of control, robotics, and autonomous systems}, vol.~4, pp. 265--293, 2021.

\bibitem{srivastava2014combined}
S.~Srivastava, E.~Fang, L.~Riano, R.~Chitnis, S.~Russell, and P.~Abbeel, ``Combined task and motion planning through an extensible planner-independent interface layer,'' in \emph{2014 IEEE international conference on robotics and automation (ICRA)}.\hskip 1em plus 0.5em minus 0.4em\relax IEEE, 2014, pp. 639--646.

\bibitem{stilman2007planning}
M.~Stilman, K.~Nishiwaki, S.~Kagami, and J.~J. Kuffner, ``Planning and executing navigation among movable obstacles,'' \emph{Advanced Robotics}, vol.~21, no.~14, pp. 1617--1634, 2007.

\bibitem{stilman2008planning}
M.~Stilman and J.~Kuffner, ``Planning among movable obstacles with artificial constraints,'' \emph{The International Journal of Robotics Research}, vol.~27, no. 11-12, pp. 1295--1307, 2008.

\bibitem{stilman2007manipulation}
M.~Stilman, J.-U. Schamburek, J.~Kuffner, and T.~Asfour, ``Manipulation planning among movable obstacles,'' in \emph{Proceedings 2007 IEEE international conference on robotics and automation}.\hskip 1em plus 0.5em minus 0.4em\relax IEEE, 2007, pp. 3327--3332.

\bibitem{krontiris2016efficiently}
A.~Krontiris and K.~E. Bekris, ``Efficiently solving general rearrangement tasks: A fast extension primitive for an incremental sampling-based planner,'' in \emph{2016 IEEE International Conference on Robotics and Automation (ICRA)}.\hskip 1em plus 0.5em minus 0.4em\relax IEEE, 2016, pp. 3924--3931.

\bibitem{havur2014geometric}
G.~Havur, G.~Ozbilgin, E.~Erdem, and V.~Patoglu, ``Geometric rearrangement of multiple movable objects on cluttered surfaces: A hybrid reasoning approach,'' in \emph{2014 IEEE International Conference on Robotics and Automation (ICRA)}.\hskip 1em plus 0.5em minus 0.4em\relax IEEE, 2014, pp. 445--452.

\bibitem{gao2022fast}
K.~Gao, D.~Lau, B.~Huang, K.~E. Bekris, and J.~Yu, ``Fast high-quality tabletop rearrangement in bounded workspace,'' in \emph{2022 International Conference on Robotics and Automation (ICRA)}.\hskip 1em plus 0.5em minus 0.4em\relax IEEE, 2022, pp. 1961--1967.

\bibitem{liu2022structformer}
W.~Liu, C.~Paxton, T.~Hermans, and D.~Fox, ``Structformer: Learning spatial structure for language-guided semantic rearrangement of novel objects,'' in \emph{2022 International Conference on Robotics and Automation (ICRA)}.\hskip 1em plus 0.5em minus 0.4em\relax IEEE, 2022, pp. 6322--6329.

\bibitem{curtis2022long}
A.~Curtis, X.~Fang, L.~P. Kaelbling, T.~Lozano-P{\'e}rez, and C.~R. Garrett, ``Long-horizon manipulation of unknown objects via task and motion planning with estimated affordances,'' in \emph{2022 International Conference on Robotics and Automation (ICRA)}.\hskip 1em plus 0.5em minus 0.4em\relax IEEE, 2022, pp. 1940--1946.

\bibitem{goodwin2022semantically}
W.~Goodwin, S.~Vaze, I.~Havoutis, and I.~Posner, ``Semantically grounded object matching for robust robotic scene rearrangement,'' in \emph{2022 International Conference on Robotics and Automation (ICRA)}.\hskip 1em plus 0.5em minus 0.4em\relax IEEE, 2022, pp. 11\,138--11\,144.

\bibitem{qureshi2021nerp}
A.~H. Qureshi, A.~Mousavian, C.~Paxton, M.~C. Yip, and D.~Fox, ``Nerp: Neural rearrangement planning for unknown objects,'' \emph{arXiv preprint arXiv:2106.01352}, 2021.

\bibitem{zeng2021transporter}
A.~Zeng, P.~Florence, J.~Tompson, S.~Welker, J.~Chien, M.~Attarian, T.~Armstrong, I.~Krasin, D.~Duong, V.~Sindhwani \emph{et~al.}, ``Transporter networks: Rearranging the visual world for robotic manipulation,'' in \emph{Conference on Robot Learning}.\hskip 1em plus 0.5em minus 0.4em\relax PMLR, 2021, pp. 726--747.

\bibitem{yuan2019end}
W.~Yuan, K.~Hang, D.~Kragic, M.~Y. Wang, and J.~A. Stork, ``End-to-end nonprehensile rearrangement with deep reinforcement learning and simulation-to-reality transfer,'' \emph{Robotics and Autonomous Systems}, vol. 119, pp. 119--134, 2019.

\bibitem{yuan2018rearrangement}
W.~Yuan, J.~A. Stork, D.~Kragic, M.~Y. Wang, and K.~Hang, ``Rearrangement with nonprehensile manipulation using deep reinforcement learning,'' in \emph{2018 IEEE International Conference on Robotics and Automation (ICRA)}.\hskip 1em plus 0.5em minus 0.4em\relax IEEE, 2018, pp. 270--277.

\bibitem{goyal2022ifor}
A.~Goyal, A.~Mousavian, C.~Paxton, Y.-W. Chao, B.~Okorn, J.~Deng, and D.~Fox, ``Ifor: Iterative flow minimization for robotic object rearrangement,'' in \emph{Proceedings of the IEEE/CVF Conference on Computer Vision and Pattern Recognition}, 2022, pp. 14\,787--14\,797.

\bibitem{wang2021uniform}
R.~Wang, K.~Gao, D.~Nakhimovich, J.~Yu, and K.~E. Bekris, ``Uniform object rearrangement: From complete monotone primitives to efficient non-monotone informed search,'' in \emph{2021 IEEE International Conference on Robotics and Automation (ICRA)}.\hskip 1em plus 0.5em minus 0.4em\relax IEEE, 2021, pp. 6621--6627.

\bibitem{wang2022lazy}
R.~Wang, K.~Gao, J.~Yu, and K.~Bekris, ``Lazy rearrangement planning in confined spaces,'' in \emph{Proceedings of the International Conference on Automated Planning and Scheduling}, vol.~32, 2022, pp. 385--393.

\bibitem{lee2021tree}
J.~Lee, C.~Nam, J.~Park, and C.~Kim, ``Tree search-based task and motion planning with prehensile and non-prehensile manipulation for obstacle rearrangement in clutter,'' in \emph{2021 IEEE International Conference on Robotics and Automation (ICRA)}.\hskip 1em plus 0.5em minus 0.4em\relax IEEE, 2021, pp. 8516--8522.

\bibitem{chaslot2008monte}
G.~Chaslot, S.~Bakkes, I.~Szita, and P.~Spronck, ``Monte-carlo tree search: A new framework for game ai,'' in \emph{Proceedings of the AAAI Conference on Artificial Intelligence and Interactive Digital Entertainment}, vol.~4, no.~1, 2008, pp. 216--217.

\bibitem{james2017analysis}
S.~James, G.~Konidaris, and B.~Rosman, ``An analysis of monte carlo tree search,'' in \emph{Proceedings of the AAAI Conference on Artificial Intelligence}, vol.~31, no.~1, 2017.

\bibitem{browne2012survey}
C.~B. Browne, E.~Powley, D.~Whitehouse, S.~M. Lucas, P.~I. Cowling, P.~Rohlfshagen, S.~Tavener, D.~Perez, S.~Samothrakis, and S.~Colton, ``A survey of monte carlo tree search methods,'' \emph{IEEE Transactions on Computational Intelligence and AI in games}, vol.~4, no.~1, pp. 1--43, 2012.

\bibitem{silver2018general}
D.~Silver, T.~Hubert, J.~Schrittwieser, I.~Antonoglou, M.~Lai, A.~Guez, M.~Lanctot, L.~Sifre, D.~Kumaran, T.~Graepel \emph{et~al.}, ``A general reinforcement learning algorithm that masters chess, shogi, and go through self-play,'' \emph{Science}, vol. 362, no. 6419, pp. 1140--1144, 2018.

\bibitem{fu2016alphago}
M.~C. Fu, ``Alphago and monte carlo tree search: the simulation optimization perspective,'' in \emph{2016 Winter Simulation Conference (WSC)}.\hskip 1em plus 0.5em minus 0.4em\relax IEEE, 2016, pp. 659--670.

\bibitem{moerland2018a0c}
T.~M. Moerland, J.~Broekens, A.~Plaat, and C.~M. Jonker, ``A0c: Alpha zero in continuous action space,'' \emph{arXiv preprint arXiv:1805.09613}, 2018.

\bibitem{labbe2020monte}
Y.~Labb{\'e}, S.~Zagoruyko, I.~Kalevatykh, I.~Laptev, J.~Carpentier, M.~Aubry, and J.~Sivic, ``Monte-carlo tree search for efficient visually guided rearrangement planning,'' \emph{IEEE Robotics and Automation Letters}, vol.~5, no.~2, pp. 3715--3722, 2020.

\bibitem{song2020multi}
H.~Song, J.~A. Haustein, W.~Yuan, K.~Hang, M.~Y. Wang, D.~Kragic, and J.~A. Stork, ``Multi-object rearrangement with monte carlo tree search: A case study on planar nonprehensile sorting,'' in \emph{2020 IEEE/RSJ International Conference on Intelligent Robots and Systems (IROS)}.\hskip 1em plus 0.5em minus 0.4em\relax IEEE, 2020, pp. 9433--9440.

\bibitem{king2017unobservable}
J.~E. King, V.~Ranganeni, and S.~S. Srinivasa, ``Unobservable monte carlo planning for nonprehensile rearrangement tasks,'' in \emph{2017 IEEE International Conference on Robotics and Automation (ICRA)}.\hskip 1em plus 0.5em minus 0.4em\relax IEEE, 2017, pp. 4681--4688.

\bibitem{chan1987iterative}
S.~K. Chan, ``An iterative general inverse kinematics solution with variable damping,'' Ph.D. dissertation, University of British Columbia, 1987.

\bibitem{lin1997collision}
M.~C. Lin, D.~Manocha, J.~Cohen, and S.~Gottschalk, ``Collision detection: Algorithms and applications,'' \emph{Algorithms for robotic motion and manipulation}, pp. 129--142, 1997.

\bibitem{krontiris2015dealing}
A.~Krontiris and K.~E. Bekris, ``Dealing with difficult instances of object rearrangement.'' in \emph{Robotics: Science and Systems}, vol. 1123, 2015.

\end{thebibliography}

\end{document}